\documentclass[letterpaper, 10pt]{article}
\usepackage{graphicx}
\usepackage{amsmath}
\usepackage{float}
\usepackage{amsfonts}
\usepackage[inner=0.75in, outer=0.75in, top=0.75in, bottom=1in, paperheight=11in, paperwidth=8.5in]{geometry}
\usepackage{algorithm}
\usepackage{algorithmicx}
\usepackage{algpseudocode}
\usepackage{epstopdf}
\usepackage{bm}
\usepackage{bbm}
\usepackage{tikz}
\usepackage{multirow}
\usepackage{amsmath}
\usetikzlibrary{bayesnet}
\usetikzlibrary{shapes.gates.logic.US,trees,positioning,arrows}
\usepackage{caption}
\usepackage{subcaption}
\usetikzlibrary{shapes,snakes}
\usetikzlibrary{trees}

\usepackage{amsthm}

\usepackage{epstopdf}
\usepackage{amssymb}
\usepackage{microtype}
\usepackage{url}
\usepackage{caption}
\usepackage{subcaption}

\usepackage[hidelinks]{hyperref}
\usepackage[capitalize]{cleveref}

\usepackage{xparse}

\begin{document}

	\title{\bf Sharing Information Between Machine Tools to Improve Surface Finish Forecasting}
	\author{D.R.\ Clarkson, L.A.\ Bull, T.A.\ Dardeno, C.T.\ Wickramarachchi, E.J.\ Cross, \\ T.J.\ Rogers, K.\ Worden, N.\ Dervilis \& A.J.\ Hughes \\ ~ \\
    Dynamics Research Group, Department of Mechanical Engineering, \\ University of Sheffield, \\ Mappin Street, Sheffield S1 3JD, UK
	}
	\date{}
	\maketitle
	\thispagestyle{empty}
	
	\section*{Abstract}
	
	At present, most surface-quality prediction methods can only perform single-task prediction \cite{wang_hidden_2002} which results in under-utilised datasets, repetitive work and increased experimental costs. To counter this, the authors propose a Bayesian hierarchical model to predict surface-roughness measurements for a turning machining process. The hierarchical model is compared to multiple independent Bayesian linear regression models to showcase the benefits of partial pooling in a machining setting with respect to prediction accuracy and uncertainty quantification.

\textbf{Keywords: Population-based Structural Health Monitoring; Bayesian Modelling, Hierarchical bayes}
	
	\section{Introduction}
	
	One of the most important measures of workpiece quality in a machining process is the surface finish, and one of the most important factors in surface finish is surface roughness. Surface roughness is a widely-used index of machined product quality \cite{ambhore_tool_2015} and a high-quality surface finish can significantly improve the fatigue strength,corrosion resistance and creep life of machined parts \cite{sharkawy_surface_2014}. The surface finish is highly important for the functional properties of parts; it has a large contribution to surface friction and the susceptibility of the part to contact wear. Additionally, the literature suggests that surface-roughness is a good indicator to estimate tool wear condition, which means accurate estimates of the surface roughness can help inform a tool condition-monitoring system \cite{abouelatta_surface_2001,venkata_rao_cutting_2013,ozel_predictive_2005}. Being able to predict surface roughness during the machining process is very valuable for manufacturers. These predictions can help inform tool replacement or inspection decision processes and reduce downtime and wasted material. 
	
	The literature showcases a wide range of modelling systems for machining features. Hidden Markov Models (HMMs), have been a popular choice \cite{wang_hidden_2002,liao_grinding_2006,wang_chmm_2012}; however, these models assume that observed values must be statistically independent of the previous sequence; this may not be the case in machining. HMM’s can lose the information between adjacent feature data which can sometimes deteriorate the recognition accuracy \cite{wang_tool_2013}.  Other researchers have used neural networks (NNs) to good effect \cite{beruvides_correlation_2014,haber_intelligent_2003,azmi_monitoring_2015}. However, NNs generally require large datasets for training which can be expensive to collect. Support Vector Machines (SVMs) are also popular but not without problems\cite{wang_force_2014}. SVMs require a selection of the kernel function and some parameters that need to be selected by trial and error; this can be tricky and leave the user with sub-optimal parameters \cite{abbasnejad_survey_2012}. Many other models have also been used such as fuzzy logic \cite{la_fe-perdomo_automatic_2019}, artificial neural network-based fuzzy inference systems \cite{shankar_prediction_2019} and chain-conditional random-field models \cite{wang_tool_2013}.
	
	Because of the natural degradation of tools during the machining process, and its effect on surface finish, tools must be replaced regularly. While each tool may be produced to the same specification and use the same materials, there will be variation within populations of tools. The variation in the physical properties of the tools is associated with variation in the behaviour between the tools; this can be an issue for standard modelling techniques. However, this variation lends itself well to a hierarchical model, a class of models that can account for variations within a population while taking advantage of the statistical similarities between them. An additional benefit of hierarchical models is their suitability to the online setting and sparse datasets; which is particularly useful for tool condition monitoring where researchers made need to make predictions as soon as the machining process has begun and with only a few data points to learn a model. Combining this with the usual benefits of Bayesian modelling (uncertainty quantification, prior information etc.) gives rise to a potentially powerful monitoring system.
	
	Hierarchical models have seen limited use in machining, Bombinski et al. highlighted the usefulness of hierarchical models by implemented a hierarchical neural network-based monitoring system with signal fusion methods\cite{jemielniak_hierarchical_2004}. Han et al. used a hierarchical structure to improve the implementation of HMMs for tool-wear estimation \cite{han_hdp-hmm_2021}. The hierarchical Dirichlet process-hidden Markov model showed greater accuracy when compared to conventional HMMs. Following the obvious advantages of hierarchical modelling for machining problems, the authors propose the use of a Bayesian hierarchical model. Specifically, a random intercepts and slopes model (also known as a mixed effects model) to predict the surface roughness during machining.

	\section{Contribution}
	
	Although Bayesian hierarchical models have seen success in other parts of engineering \cite{bull_hierarchical_2023,papadimas_hierarchical_2021,francesco_decision-theoretic_2021},the benefits of these models have not yet reached machining and tool health monitoring. In this paper, the authors propose a random intercepts and slopes model to show the modelling improvements of hierarchical models specifically, for sparse datasets in machining.

	\section{The Data}\label{sec:RBSHM}

\begin{figure}[!h] 
	  \centering{\includegraphics[scale=0.75]{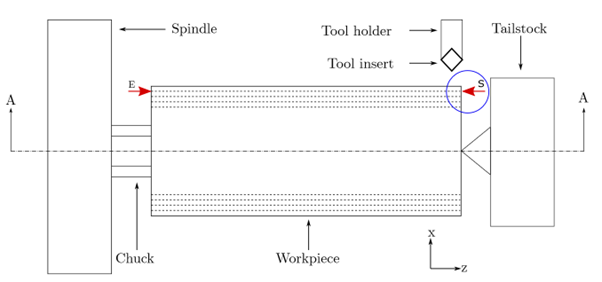}}
		\caption{Schematic showing the experimental set up used for data acquisition  \cite{wickramarachchi_automated_2019}.}
	\label{fig:experiment}%
	\end{figure}

	The dataset analysed in this paper is from the turning process shown in Figure \ref{fig:experiment}. The workpiece is rotated around the \emph{z-axis} and the tool makes four passes along the workpiece. Each pass starts at point S and ends at point E. After four passes, the tool is inspected, and measurements are taken of the workpiece and tool. The four passes and measurements are repeated until tool failure.  For full details of the experiment refer to Wickramarachchi \cite{wickramarachchi_automated_2019}.

\begin{figure}[!h] 
	  \centering{\includegraphics[scale=0.8]{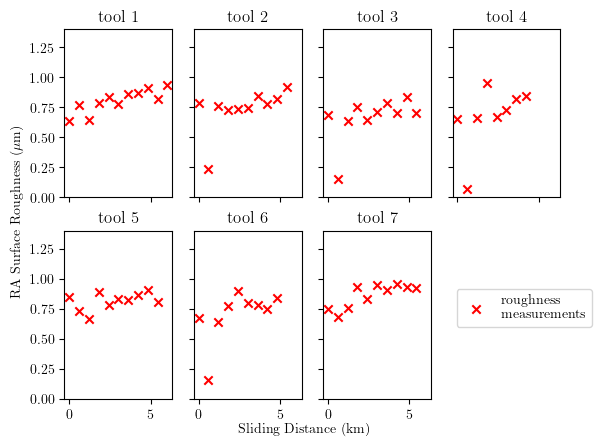}}
		\caption{Experimental surface roughness measurements.}
	\label{fig:data}%
	\end{figure}

	The data to be analysed in this paper consists of the workpiece surface roughness measurements from seven repeats of the experiment detailed above. After each experiment the tool is replaced with a fresh tool. This data can be seen in Figure \ref{fig:data}. The plots show arithmetic mean ($R_{a}$) surface roughness measurements against sliding distance. Sliding distance is how far the tool has traveled along the work piece, it is effectively a measure of how long the tool has been machining for. $R_{a}$ surface roughness measures the deviation of a surface from a theoretical centre line \cite{stahl_analytical_2011}.
	
	\section{The Hierarchical Model}

	The explanation follows the description provided by Bull et al. \cite{bull_hierarchical_2023}. Consider machining data, recorded from a population of $K$ similar tools. The population data can be denoted,
	\begin{equation}\label{eq:data}
	  \left\{\mathbf{x}_k, \mathbf{y}_k\right\}_{k=1}^K=\left\{\left\{x_{i k}, y_{i k}\right\}_{i=1}^{N_k}\right\}_{k=1}^K
	\end{equation}
	where $y_{k}$ is a target response vector for inputs $x_{k}$ and $\{x_{ik},y_{ik}\}$ are the $i^{\text{th}}$ pair of observations in group $k$. There are $N_{k}$ observations in each group and thus $\Sigma^{K}_{k=1}N_k$ observations in total. The aim is to learn a set of $K$ predictors related to the regression task. This paper focuses on regression, where the tasks satisfy,
	\begin{equation}\label{eq:regression}
	  \left\{y_{i k}=f_k\left(x_{i k}\right)\right\}_{k=1}^K
	\end{equation}
	and the output $y_{ik}$ is determined by evaluating one of $K$ latent functions. For the case of linear regression, the mapping is denoted by,
	\begin{equation}\label{eq:linearregression}
	  f_k\left(x_{i k}\right)=m_k\left(x_{i k}\right)+c_{ k}+\epsilon_{k}
	\end{equation}
	where $m_{k}$ is the tool-specific gradient of the roughness, $c_{k}$ is the tool-specific intercept and $\epsilon_{k}$ is the tool-specific noise. Where the noise is assumed to be $\epsilon_{k} {\sim}\mathrm{Cauchy}\left(0,{\gamma}_{k}\right)$  Together they form the set of $K$ predictors,
	\begin{equation}\label{eq:predictors}
	  \left\{{c}_k, {m}_k,{\epsilon}_{k}\right\}_{k=1}^K
	\end{equation}

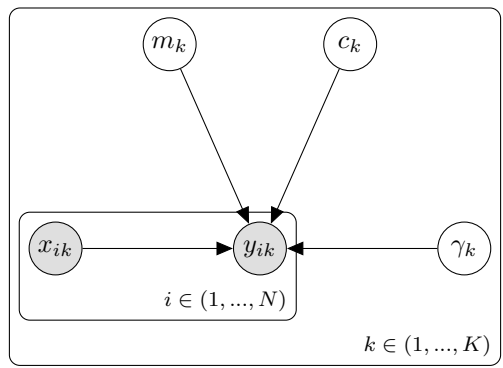
\begin{figure}[!b]
	  \centering
	  \begin{tikzpicture}
	    \node[obs]                               (y) {$y_{ik}$};
	    \node[obs, left=2cm of y]               (x) {$x_{ik}$};
	    \node[latent, above=2cm of y, xshift=-1.2cm] (m) {$m_{k}$};
	    \node[latent, above=2cm of y, xshift=1.2cm] (c) {$c_{k}$};
	    \node[latent, right=2cm of y]  (noise) {$\gamma_{k}$};

	    \edge {c,m,x} {y} ; %
	    \edge {noise} {y} ; %

	    \plate {N} {(x)(y)} {$i\in(1,...,N)$} ;
	    \plate {k} {(x)(y)(m)(c)(noise)(N.north west)(N.south east)} {$k\in(1,...,K)$} ;

	  \end{tikzpicture}
	  \caption{A graphical model representing the independent model.}
	  \label{fig:independent_model}
	\end{figure}

	In this paper, comparisons will be made between a \emph{hierarchical} model, where the mapping $f_k$ \emph{is} assumed to be correlated between tools and an \emph{independent} model where correlation is not assumed.
	For the independent model, the slope and intercept of each tool are learned independently. A graphical model depicting the independent model can be seen in Figure \ref{fig:independent_model} .

	Since the mapping $f_k$ is assumed to be correlated between tools for the hierarchical model, the model should be improved by learning the parameters in a joint inference over the whole population. The hierarchical model learns a global distribution over the tools and assumes the gradient and intercept associated with each tool is a sample from this global distribution. In practice, while a tool that has been in use for some extended amount of time may have rich historical data, newly-replaced tools will have limited training data. In this setting, learning separate, independent models for each group will lead to unreliable predictions. On the other hand, a single regression of all the data (complete pooling) will result in poor generalisation. Instead, hierarchical models can be used to learn separate models for each group while encouraging tool-specific parameters to be correlated (partial pooling). The likelihood for the model is,
	\begin{equation} \label{eq:likelihood}
	  \left\{{y}_{ik}\right\}_{k=1}^K {\sim} \mathrm{Cauchy}\left({m}_{k}\cdot{x}_{ik}+{c}_{k},{\gamma}_{k}\right)
	\end{equation}

	Following the Bayesian methodology, one can set prior distributions over the slope and intercept for the groups,
	\begin{equation} \label{eq:m_k}
	  \left\{{m}_k\right\}_{k=1}^K {\sim} \mathrm{Cauchy}\left({\mu}_m,{\sigma}_{m}\right)
	\end{equation}
	\begin{equation} \label{eq:mu_m}
	  {\mu}_m \sim \mathrm{Cauchy}\left({\bar{\mu}}_{m},s_{\mu_{m}}\right)
	\end{equation}
	\begin{equation} \label{eq:sig_m}
	  {\sigma}_m \sim \mathrm{HalfCauchy}\left(0,s_{\sigma_{m}}\right)
	\end{equation}
	where the slopes are $\mathrm{Cauchy}$ distibuted, with mean ${\mu}_m$ and standard deviation $\sigma_{m}$. Equation $\left(\ref{eq:mu_m}\right)$ shows the prior expectation of the slopes is also $\mathrm{Cauchy}$ distributed with mean $\bar{\mu}_{m}=0$  and standard deviation $s_{\mu_{m}}=1$. Equation $\left(\ref{eq:sig_m}\right)$ shows that the prior deviation of the slope is $\mathrm{HalfCauchy}$ distributed with scale parameter $s_{\sigma_{m}}=1$.
	\begin{equation} \label{c_k}
	  \left\{{c}_k\right\}_{k=1}^K {\sim} \mathrm{Cauchy}\left({\mu}_c,{\sigma}_{c}\right)
	\end{equation}
	\begin{equation} \label{eq:mu_c}
	  {\mu}_c \sim \mathrm{Cauchy}\left({\bar{\mu}}_{c},s_{\mu_{c}}\right)
	\end{equation}
	\begin{equation} \label{eq:sig_c}
	  {\sigma}_c \sim \mathrm{HalfCauchy}\left(0,s_{\sigma_{c}}\right)
	\end{equation}
	where the intercepts are $\mathrm{Cauchy}$ distributed, with mean ${\mu}_c$ and standard deviation $\sigma_{c}$. Equation $\left(\ref{eq:mu_c}\right)$ shows the prior expectation of the intercepts is also $\mathrm{Cauchy}$ distributed with mean $\bar{\mu_{c}}=0$ and standard deviation $s_{\mu_{c}=1}$. Equation $\left(\ref{eq:sig_c}\right)$ shows that the prior deviation of the intercept is $\mathrm{0,HalfCauchy}$ distributed with scale parameter $s_{\sigma_{c}}=1$.

	\begin{equation} \label{eq:noise}
	  \left\{\mathbf{\gamma}_k\right\}_{k=1}^K {\sim} \mathrm{HalfCauchy}\left(\gamma\right)
	\end{equation}

	\begin{equation} \label{eq:sig_noise}
	  \mathbf{\gamma} {\sim} \mathrm{HalfCauchy}\left(0,s_{\gamma}\right)
	\end{equation}

\begin{figure}[!h]
	  \centering
	  \begin{tikzpicture}
	    \node[obs]                              (y) {$y_{ik}$};
	    \node[obs, left=2cm of y]               (x) {$x_{ik}$};
	    \node[latent, above=1.5cm of y, xshift=-1.5cm] (m) {$m_{k}$};
	    \node[latent, above=1.5cm of y, xshift=1.5cm] (c) {$c_{k}$};
	    \node[latent, right=1.5cm of y]  (noise) {$\gamma_{k}$};
	    \node[latent, above=1.5cm of m, xshift=0.8cm] (mu_m) {$\mu_{m}$};
	    \node[latent, above=1.5cm of m, xshift=-0.8cm] (sig_m) {$\sigma_{m}$};
	    \node[latent, above=1.5cm of c, xshift=0.8cm] (mu_c) {$\mu_{c}$};
	    \node[latent, above=1.5cm of c, xshift=-0.8cm] (sig_c) {$\sigma_{c}$};
	    \node[latent, above=1cm of noise, xshift=1.5cm] (hypernoise) {$\gamma$};

	    \edge {c,m,x} {y} ; %
	    \edge {noise} {y} ; %
	    \edge {mu_m,sig_m} {m} ; %
	    \edge {mu_c,sig_c} {c} ; %
	    \edge {hypernoise} {noise} ; %

	    \plate {N} {(x)(y)} {$i\in(1,..,N)$} ;
	    \plate {k} {(x)(y)(m)(c)(noise)(N.north west)(N.south east)} {$k\in(1,..,K)$} ;

	  \end{tikzpicture}
	  \caption{A graphical model representing the hierarchical model with partial pooling.}
	  \label{fig:hierarchical_model}
	\end{figure}
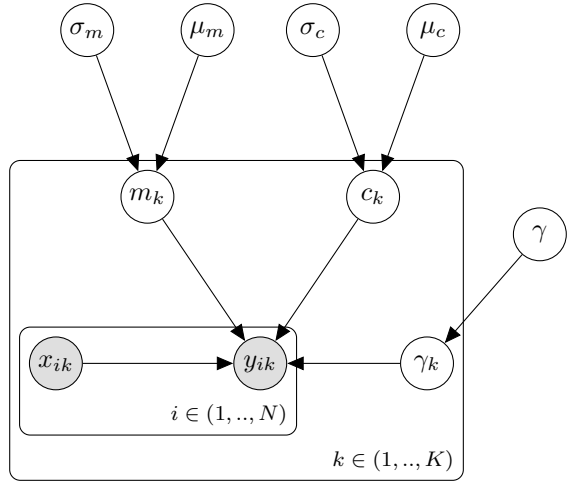

	Finally, the variance of $y_{k}$, $\gamma_{k}$, is $\mathrm{HalfCauchy}$ distributed. Equation $\left(\ref{eq:sig_noise}\right)$ shows that the prior deviation of $\gamma_{k}$ is $\mathbf{\gamma}$ which is half $\mathrm{HalfCauchy}$ distributed with scale $s_{\gamma}$. In this paper, $s_{\gamma}=$ 1. As recommended by Gelman et al. \cite{gelman_prior_2006}, Cauchy distributions are used. Their heavy tails bring a robustness against outliers to the model, as well as efficiency during the inference and sampling process. A graphical model depicting the hierarchical structure can be seen in Figure \ref{fig:hierarchical_model}.

	\section{Results}

	The hierarchical and independent models will now be compared. For Tools 1-5, every measurement is given to the models for training. However, for Tools 6 and 7, the training set is restricted to only the first 5 roughness measurements. This emulates a scenario in which Tools 1-5 are no longer in use, and have completed the full tool life cycle, while Tools 6 and 7 are new tools with limited measurements. What one would expect to see is the independent model making accurate predictions for Tools 1-5, where there is sufficient data for the model to learn; but for Tools 6 and 7, the independent model is expected to struggle. Since this model is computing independent regressions for each tool, for tools with a smaller number of measurements, the model will be uncertain in its predictions due to the lack of data. In contrast, the hierarchical model will be able to use what it has learnt from the previous tools to reduce uncertainty in predictions. An additional benefit for the hierarchical model is the in-built robustness, because the model has seen Tools 1-5 before and remembers this data via updates to the global distributions of the gradient and slope, it is resistant to new, unrepresentative, data. Another way to visualise this is that outliers are more diluted since the previous observations count as extra data for this new tool.

\begin{figure}[!h] 
	  \centering{\includegraphics[scale=0.85]{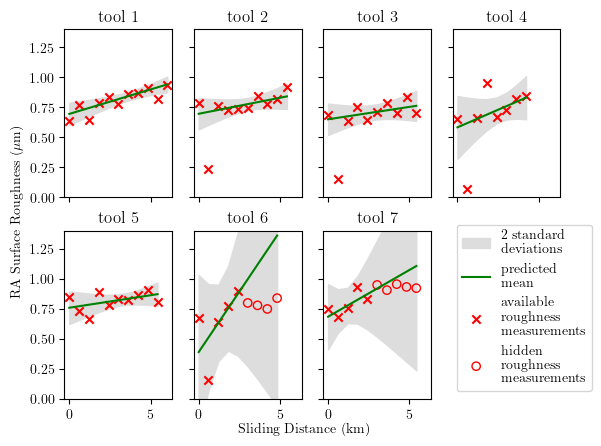}}
	    \caption{The output from the independent model. The y-axes of all figures in this paper have been limited between 0-1.4 $\mu m$ for ease of comparison.}
	\label{fig:independent_hidden}%
	\end{figure}

	The predictions of the independent model when trained on the roughness measurements can be seen in Figure \ref{fig:independent_hidden}. The red crosses are the data the model has been trained on, while the red circles are the measurements the model cannot see, the green line is the predicted mean and the grey area is two standard deviations from the mean. Under data-rich conditions, Tools 1-5, the independent model fits well to the data. The model seems to fit a good estimate of the mean roughness but the standard deviations are large in some instances. For example, for Tool 4 it can be seen that two of the data points are far from the mean and cause large uncertainty in the model. Large uncertainties could cause problems in industry. For example, a simple tool condition-monitoring system may have some acceptable surface roughness, and when the roughness measurements surpass this value the tool must be replaced. In this scenario having uncertainties this large could cause false triggers of tool replacements; this will waste time and money for manufacturers.

	\begin{figure}[!h] 
	  \centering{\includegraphics[scale=0.85]{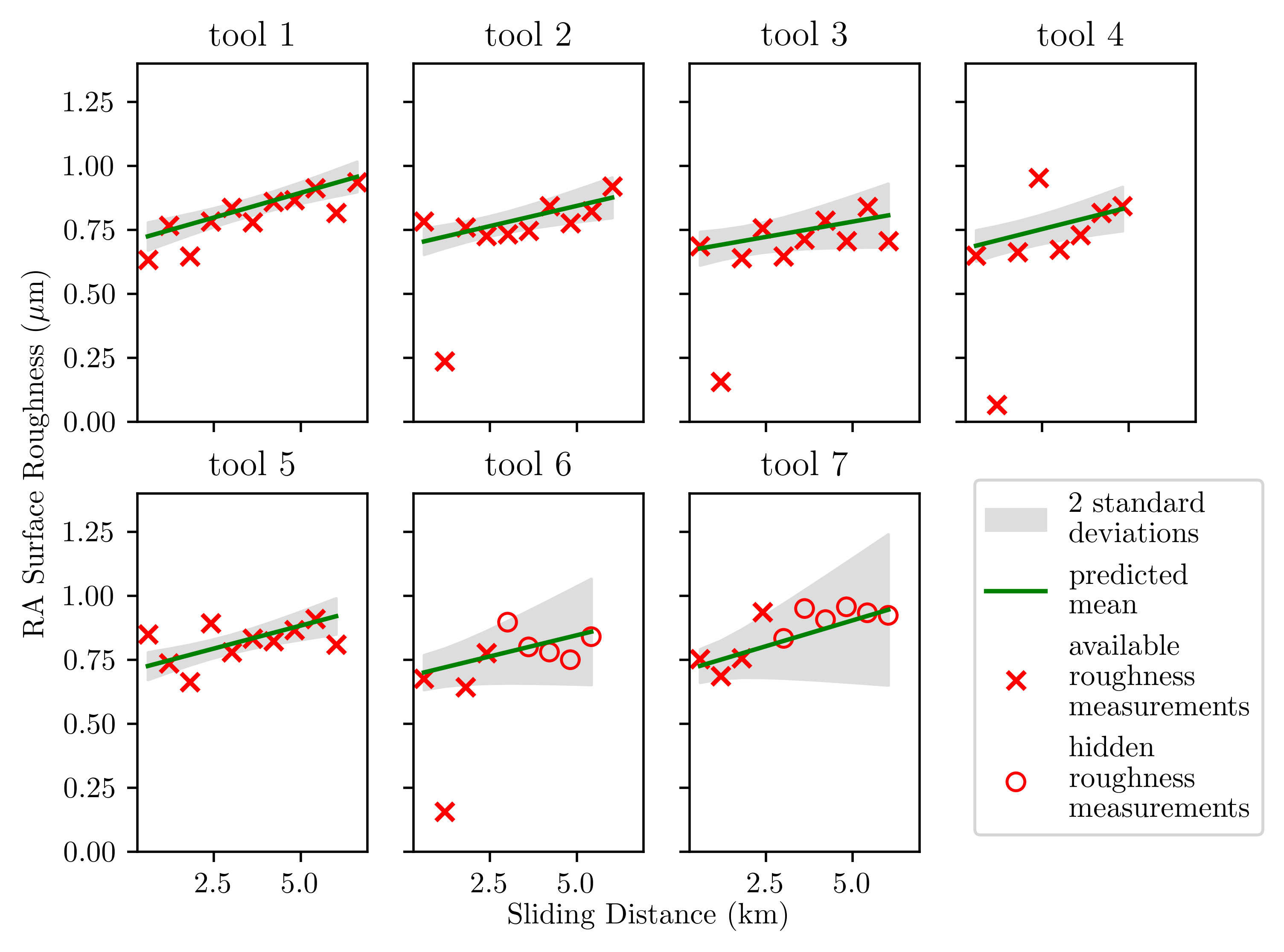}}
	    \caption{The output from the hierarchical model.}
	\label{fig:hierarchical_hidden}%
	\end{figure}

	For Tools 6 and 7, with so few data points, the standard deviation of the independent model suffers. The model has over-estimates the variance from to the available data and does a poor job of predicting the hidden measurements. There is such large levels of uncertainty that the mean predictions are effectively meaningless. 
	
	Compare this situation to the hierarchical model which can be seen in Figure \ref{fig:hierarchical_hidden}, again, for Tools 1-5 the model fits the data well and the predicted means look sensible. Where the models differ is in the standard deviations. The hierarchical model is more confident in its predictions, as can be seen by the smaller grey area. This model is less likely to trigger an unnecessary replacement of the tool, increasing the efficiency of the manufacturing process.

	Where the differences between the models is highlighted best in the data poor scenarios, Tool 6-7. As expected, the hierarchical model performs better. The mean predictions do a good job of predicting the hidden measurements and the uncertainty in these predictions is smaller. The hierarchical model can draw on the statistical strength of the measurements from other tools which means that it is less prone to over-estimating the variance in the data-sparse setting.
	
	\section{Conclusion}

	In this paper a Bayesian hierarchical model was used to predict workpiece surface roughness as a function of sliding distance. A clear benefit to the model was shown by comparisons to a set of independent linear regressions. The improved predictions and uncertainty quantification is useful when making predictions for a new tool without a rich history of data. The use of Bayesian hierarchical models can help improve decision-making processes and reduce costs involved in machining. Looking forward, the hierarchical model will be used to compute risk in an active learning framework and inform a decision making process for inspecting the machining tool.

	\section*{Acknowledgements}
	The authors would like to gratefully acknowledge the support of the UK Engineering and Physical Sciences Research Council (EPSRC) via grant references EP/W005816/1. For the purposes of open access, the authors have applied a Creative Commons Attribution (CC BY) license to any Author Accepted Manuscript version arising.
	
	\bibliographystyle{iwshm}
	\bibliography{IWSHM-my_references6}

\end{document}